\documentclass[letterpaper,10pt,conference]{IEEEtran}
\IEEEoverridecommandlockouts
\usepackage[T1]{fontenc}
\usepackage{amsmath,amssymb}
\usepackage{booktabs}
\usepackage{graphicx}
\usepackage{float}
\usepackage{multirow}
\usepackage{xcolor}
\usepackage{array}
\usepackage{tabularx}
\usepackage[hidelinks]{hyperref}
\usepackage{cite}
\usepackage{placeins}
\usepackage{url}
\usepackage{tikz}
\usetikzlibrary{arrows.meta,positioning,fit,calc}

\newcommand{\topk}{\ensuremath{\operatorname{Top\!\text{-}\!K}}}
\newcommand{\routepool}{\mathcal{P}^{\mathrm{route}}}
\newcommand{\frontendpool}{\mathcal{P}^{\mathrm{raw}}}
\newcommand{\methodname}{\textsc{Route-MHT}}

\title{\bf\Large%
Route-MHT: \\
Multimodal Transformer Guardrails for Thermal Visual Place Recognition
}
\author{Zhiyuan Lu ~~~~~~~~~ Kanji Tanaka}
\hypersetup{pdfauthor={Anonymous Authors},pdftitle={Route-MHT: Multimodal Transformer Guardrails for Thermal Visual Place Recognition}}
\begin{document}
\maketitle

\begin{abstract}
Strong mapped-region thermal visual place recognition (VPR) does not ensure
safe rejection of unmapped queries. We identify and quantify this gap in
AnyThermal: high Map-In retrieval accuracy coexists with confident false loop
closures in Map-Out. We address it with \methodname{}, a multimodal transformer
guardrail. Causal motion forms a route-local candidate pool beyond the
frontend's Top-$K$. Closed-form $SE(2)$ SVD verifies candidates, while frozen
visual features, nine-dimensional SVD residuals, and motion proxies enter a
masked multi-head transformer (MHT). Their interactions yield a contextual
confidence correction to reject unsupported matches without altering geometric
pose alignment.

We collect an indoor thermal dataset with a physical mobile robot (Dataset-A),
forming five same-day/cross-day map--query pairs; the protected interface reaches macro
R@1@5/10~m of $.610/.861$. Dataset-B comprises 20 map-hole scenarios derived
from public STheReO--KAIST recordings. Across five scenario-held-out folds
and three seeds, \methodname{} reduces FPR from $.116$ for the SVD baseline to $.061$ (paired 95\% CI
$[-.100,-.014]$), while improving AUC from $.945$ to $.967$ and recall from
$.884$ to $.914$. Public benchmark transfer checks on STheReO--KAIST, MS2,
and IRSLAM--KRI extend the evaluation under their released
metric or route-progress protocols.
\end{abstract}

\section{Introduction}
Thermal visual place recognition (VPR) supports inspection robots, warehouse
fleets, and long-horizon patrol where visible-light imagery is unreliable.
Modern frozen descriptors improve cross-condition retrieval
\cite{keetha2024anyloc,izquierdo2024salad,maheshwari2026anythermal}, but they
still solve a closed-set ranking problem: every query receives a database
match. A robot can therefore be confidently wrong after a route change, map
update, or partial remap. Preventing that false loop is a navigation-safety
problem, not merely another Recall@1 improvement. Figure~\ref{fig:first-page-teaser}
contrasts supported-map retrieval with unmapped-query rejection.

\begin{figure}[!t]
\setcounter{figure}{0}
\centering
\includegraphics[width=\columnwidth]{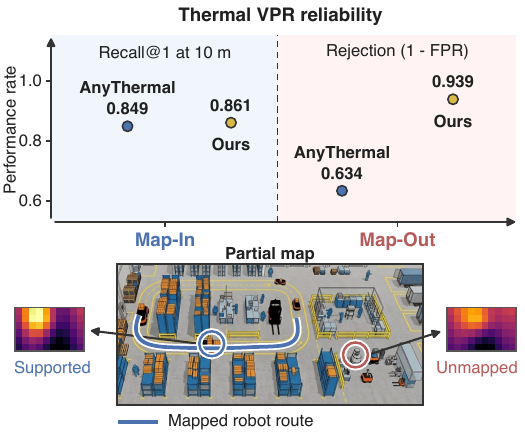}
\caption{\textbf{Map-In versus Map-Out trust decisions.}
Left: AnyThermal achieves $.849$ macro Recall@1 at 10~m on five Dataset-A
pairs; the protected proposed interface achieves $.861$. Right: on 20 Dataset-B
map-hole scenarios, rejection increases from $.634$ (AnyThermal) to $.939$
(ours). The horizontal categories are Map-In and Map-Out; the vertical
scale reports their distinct metrics. In the illustrative warehouse scene,
blue marks the mapped robot route; the red-ringed robot lies outside its
coverage. Yellow lines are floor markings. The scene is an AI-generated
mockup, not an experimental photograph; thermal insets are schematic.}
\label{fig:first-page-teaser}
\end{figure}

Strong closed-set thermal retrieval does not by itself protect open-world
navigation from unsupported matches. In this work, we make the first
systematic Map-In/Map-Out characterization of AnyThermal (ICRA 2026)
\cite{maheshwari2026anythermal}, which achieves macro Recall@1 of $.849$ at 10~m
on our five mapped-region pairs (Map-In, Figure~\ref{fig:first-page-teaser}).
Yet its calibrated rejection rate is only $.634$ on our 20 unmapped-region
scenarios (Map-Out), exposing robots to false-positive loop closures.
These distinct metrics reveal a gap between retrieval accuracy and match
support. To address this safety gap, we propose a dedicated backend guardrail
that retains AnyThermal as the retrieval frontend while systematically auditing
whether the map supports a match.

Addressing this risk requires separating coverage from selection, and map
support from match correctness. A selector cannot recover a location absent
from its pool, so we decompose recall as $R=C\cdot E$, with coverage $C$ and
selection efficiency $E$. A nearby mapped location also does not guarantee a
correct selection. Controlled map removal isolates unsupported-match acceptance,
motivating candidate re-query followed by calibrated validity auditing.

We propose \methodname{} (Route-based Multi-Head Transformer), a causal
multimodal trust layer with three explicit roles (Figure~\ref{fig:pipeline}).
A sequence-supported route mode re-queries a bounded map neighborhood beyond
the frozen Top-100 candidate set. A deterministic Singular Value Decomposition
(SVD)-based $SE(2)$ estimator verifies this pool. Multi-head attention then
combines visual relations, SVD residuals, and motion to accept or reject the
match. A bounded correction preserves baseline confidence as a safety prior;
the learned module audits validity without replacing the exact SVD pose solver.

Our contributions are:
\begin{itemize}
    \item the first systematic Map-In/Map-Out characterization of AnyThermal,
    exposing the gap between closed-set retrieval and unsupported-match rejection;
    \item a multimodal transformer guardrail with causal route-local re-query
    that combines visual, geometric, and motion attributes while preserving
    the exact SVD solver; and
    \item thermal data collected with a physical mobile robot for five-pair
    localization evaluation, complemented by a public-data 20-scenario map-hole
    protocol with disjoint scenario splits, three seeds, and paired intervals.
\end{itemize}

An anonymized reproducibility artifact with frozen scenario definitions and
integrity hashes is submitted separately.
\section{Related Work}
Sequence VPR uses temporal structure to reduce appearance ambiguity, from
SeqSLAM \cite{milford2012seqslam} to learned sequential descriptors
\cite{mereu2022sequential,lowry2016visual,schubert2021multi}. These methods motivate causal history, but a
reranker cannot retrieve a map location excluded from its input list. We
therefore compare against a descriptor-adapted SeqSLAM baseline and separately
measure the effect of re-querying the full map around a causal route mode.

Modern global descriptors and pairwise verification improve closed-set retrieval.
Examples include NetVLAD
\cite{arandjelovic2016netvlad}, Patch-NetVLAD \cite{roxas2021patchnetvlad},
CosPlace \cite{lettry2022cosplace}, MixVPR \cite{ali2022mixvpr}, and Pair-VPR
\cite{hausler2025pairvpr}
with global aggregation or learned verification. Route-MHT addresses a different failure event:
whether a retrieved anchor is supported by the current map and trajectory.
Accordingly, we retain the frozen thermal descriptor as a candidate source and
evaluate map-out false acceptance separately from closed-set recall. This
distinction makes the comparison complementary rather than a claim that a
guardrail should replace a high-performing retrieval frontend.

Geometric verification provides evidence distinct from image similarity.
\methodname{} uses closed-form weighted $SE(2)$ Procrustes alignment
\cite{huber1964,grisetti2010tutorial,grisetti2011tutorial} to score candidates and expose an auditable residual. Unlike
complete multi-hypothesis localization or graph-SLAM
\cite{dellaert1999mcl,dellaert2017factor}, this backend does not maintain a
global posterior or optimize a factor graph; it accepts or rejects a bounded
VPR match.

Out-of-distribution (OOD) and uncertainty methods estimate whether a model
output should be trusted
\cite{hendrycks2017baseline,liang2018odin,zaffar2024uncertainty,guo2017calibration,lakshminarayanan2017simple}. In VPR,
however, selected-index correctness and map support require different labels.
Our retained-map-hole protocol removes map segments before retrieval and
measures false acceptance at a threshold selected only on the calibration
split. This controlled intervention is reported separately from closed-set
localization accuracy.
\section{Method}
\begin{figure*}[!t]
\centering
\includegraphics[width=17cm]{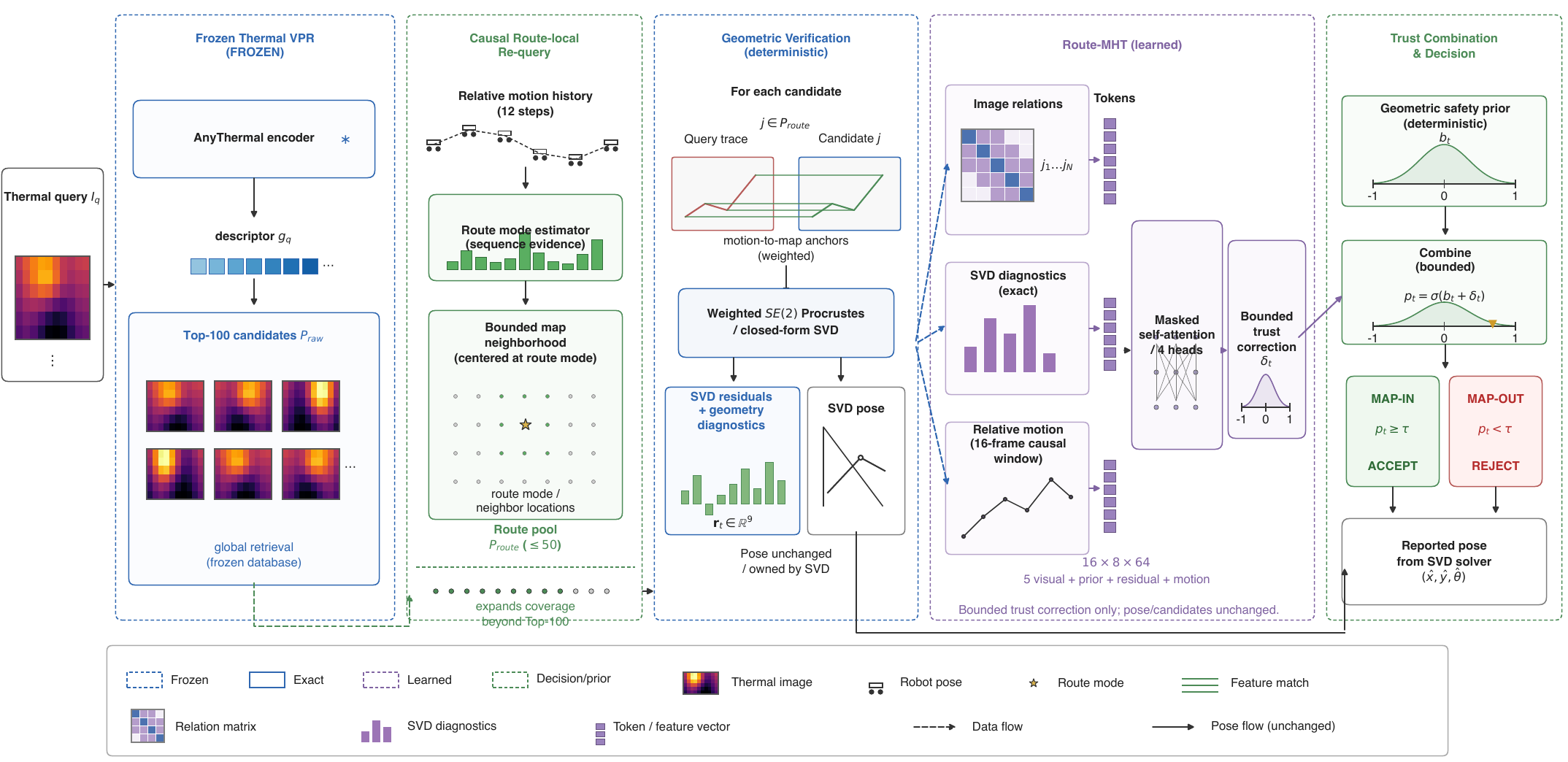}
\caption{\textbf{Separate geometric estimation and learned match acceptance.}
Frozen retrieval seeds a route-local re-query using 12 steps of motion history.
Exact SVD aligns the relative-motion trace with candidate map anchors.
Image relations, SVD diagnostics, and motion enter a 16-frame causal attention
window. The learned head adds a bounded logit correction to the geometric
prior; a frozen threshold gates the match. Only accepted matches pass the SVD
output; rejection adds no loop constraint. Schematics illustrate data flow,
not image-keypoint correspondences. Absolute query coordinates and correctness
labels are excluded from runtime tokens. The tensor shape denotes
16 time steps, eight token groups, and width 64; $b$ is the scalar geometric
prior, $\delta$ the bounded correction, and the decision threshold is frozen
on calibration data.}
\label{fig:pipeline}
\end{figure*}

Figure~\ref{fig:pipeline} summarizes the pipeline: causal route-local retrieval,
exact SVD verification, and multimodal match acceptance.

\subsection{Online Information Boundary}
The backend uses only map data and causally available query observations.
Let $\mathcal{M}=\{(\mathbf{x}_j,\mathbf{f}_j)\}_{j=1}^M$ denote map
coordinates and frozen map descriptors. At query time $t$, the frontend yields
a descriptor $\mathbf{g}_t$, similarities $s_t(j)$, and a raw candidate list
\begin{equation}
\frontendpool_t=\topk_{j\in\{1,\ldots,M\}}s_t(j), \qquad K=100.
\end{equation}
The online system additionally receives relative forward motion and turn
increments. Query absolute position, candidate correctness, map-support labels,
and any oracle reset are prohibited. Query coordinates do not enter the runtime
array directly; the provenance of the packaged relative-motion stream is
documented in Section~\ref{sec:evaluation-protocol}.

\subsection{Theoretical View: Coverage, Efficiency, and Causal Information Flow}
\label{sec:theory}
Final Recall@1 factorizes into candidate coverage and selection efficiency.
The working set must contain a correct location, and the selector must pick it:
\begin{equation}
R_m(r)=\underbrace{C_m(r)}_{\text{pool coverage}}
\cdot\underbrace{E_m(r)}_{\text{selection efficiency}},
\label{eq:rce}
\end{equation}
with $C_m$, $R_m$, and $E_m$ defined in Section~\ref{sec:evaluation-protocol}. A margin
gate on raw Top-1 similarity can only preserve $R$ when the frontend is already
correct; it cannot increase $C$ because it never changes the selected map index.
Route-local re-query is therefore a \textbf{causal information-flow} change: it
replaces $\frontendpool_t$ with $\routepool_t$ before any geometric selector runs.

Candidate selection and map-support auditing share the same causal information
but optimize different targets. Let $\mathcal{I}_t$ denote all information available at time $t$ under the
online boundary (current descriptor, raw similarities, causal relative motion,
and map geometry). The route mode $\hat{j}^{\mathrm{route}}_t$ is a function
of $\mathcal{I}_t$ only; the pool $\routepool_t$ is a deterministic map
neighborhood of that mode. SVD baseline and appearance selectors are measurable maps
from $(\mathcal{I}_t,\routepool_t)$ to a discrete map index. Map-out
guardrails use the same $\mathcal{I}_t$ but optimize a different target:
operating-point false-positive rate under retained-map-hole labels, which is
orthogonal to closed-set $R_m(r)$ when the hole is not in the raw Top-100.
This decomposition explains why the paper reports both VPR tables and
guardrail tables without treating AUC on natural strict-5~m support as a
deployment guarantee.

\subsection{Causal Sequence Mode and Route-Local Candidate Re-query}
The evaluated route mode is selected by causal sequence evidence, not by the
diagnostic posterior reported as a separate baseline. For each current raw
candidate $j\in\frontendpool_t$, we trace its route arclength $a_j$ backward
through a $W=12$-step history using cumulative absolute forward motion
$d_{t,\ell}=\sum_{r=t-\ell+1}^{t}|u_r|$. Its evidence is
\begin{align}
\mathcal{N}_{t,\ell}(j) & =
\left\{k:\left|a_k-(a_j-d_{t,\ell})\right|\leq\epsilon\right\}, \\
 b_{t,\ell}(j) & = \max_{k\in\mathcal{N}_{t,\ell}(j)}s_{t-\ell}(k), \\
 e_t(j) & = (1+h_t)^{-1}\left[s_t(j)+\sum_{\ell=1}^{h_t}b_{t,\ell}(j)\right], \\
 h_t & = \min(W,t),\qquad \epsilon=4\text{ m}.
\label{eq:sequence-evidence}
\end{align}
where the maximization scans the known map arclength neighborhood in each past
frame. The causal route mode is
$\hat{j}^{\mathrm{route}}_t=\arg\max_{j\in\frontendpool_t}e_t(j)$; it uses no
future image, query coordinate, or correctness label.

Route-local re-query can recover candidates outside the frozen retrieval list.
The route mode defines the pool
\begin{equation}
\routepool_t=\left\{j:\left\|\mathbf{x}_j-
\mathbf{x}_{\hat{j}^{\mathrm{route}}_t}\right\|_2\leq 12\text{ m}\right\},
\label{eq:pool}
\end{equation}
from which we retain at most 50 map locations by current-frame appearance.
Equation~\eqref{eq:pool} re-queries the known map around the selected route,
rather than merely reordering $\frontendpool_t$. The pool can be smaller than
the raw Top-100, and it can also include locations absent from it; this is why
coverage must be reported rather than inferred from the nominal budget.

\subsection{Same-Pool Selectors and Confidence Preservation}
The appearance-only control isolates selection within the same route-local pool.
Pairing appearance and SVD on this pool tests geometric refinement without
changing candidate coverage. The appearance selector returns
\begin{equation}
\hat{j}^{\mathrm{app}}_t=\arg\max_{j\in\routepool_t}s_t(j).
\end{equation}

\subsection{Route-MHT Multimodal Map-Out Guardrail}
\label{sec:route-mht}
Route-MHT separates \emph{geometric estimation} from \emph{map-support
judgment}. Its route-local SVD branch remains the only component that estimates the
candidate pose and map index. The multimodal MHT guardrail receives three token
streams after SVD: the frozen thermal
image descriptor, the numerical SVD diagnostics, and the causal motion proxy.
The MHT predicts only an in-map confidence correction; it never recomputes the
SVD alignment or replaces the selected index. This makes the learned module a
testable multimodal decision layer rather than a post-hoc scalar evaluator.

\paragraph{SVD geometric verification.}
Each candidate $j\in\routepool_t$ is scored by a causal weighted $SE(2)$
Procrustes fit between the 12-frame relative-motion trace and map anchors traced
backward from that branch. Branch cost $\mathcal{C}_t(j)$ is a 75\%-trimmed
weighted Huber geometric residual ($\delta=8$~m) \cite{huber1964} plus twice a
weighted appearance dissimilarity. The geometric output remains
\begin{equation}
\hat{j}^{\mathrm{SVD}}_t=
\arg\min_{j\in\routepool_t}\mathcal{C}_t(j),
\label{eq:svd-selector}
\end{equation}
with rotation and translation computed internally by the closed-form SVD solve.
No learned model recomputes this alignment or substitutes a learned pose.

SVD diagnostics expose geometric fit quality and observability to the guardrail.
The winning branch returns
\begin{equation}
\begin{aligned}
\mathbf r_t=\bigl[&-\mathcal C_t(\hat{j}^{\mathrm{SVD}}_t),\ \bar e_t,
 \bar e^{\rm trim}_t,\ C^{\rm geom}_t,\ C^{\rm fit}_t,\\[-1mm]
 &\sigma_{1,t},\ \sigma_{2,t},\ \kappa_t,\ q^{\rm geom}_t\bigr]\in\mathbb R^9,
\end{aligned}
\label{eq:svd-residual-vector}
\end{equation}
where the remaining terms are evaluated on the same winning branch. The vector
contains the primary negative geometric baseline cost, mean and trimmed residuals, geometric
and fit costs, the two singular values, anisotropy $\kappa_t$, and geometric
confidence $q^{\rm geom}_t$. These are physical outputs of the SVD branch, not
features reconstructed from ground-truth query pose.

\paragraph{Causal confidence and rejection.}
Calibrated causal confidence gates map support without changing the SVD estimate.
The relative-motion stream forms
\begin{equation}
\mathbf u_t=[\Delta f_t,\ \operatorname{wrap}(\Delta\psi_t),\
 |\operatorname{wrap}(\Delta\psi_t)|,\ \Delta f_t/\Delta t_t]
 \in\mathbb R^4,
\label{eq:motion-token}
\end{equation}
where timestamps make speed distinct from displacement and angle differences
are wrapped before normalization. Let $c_t$ denote the standardized winning
geometric baseline cost and let $q_t$ denote a causal trajectory-support score formed from the
current score, the exponentially weighted moving average (EWMA) appearance
score, and the recent SVD residuals. The
operating threshold $\tau_{\rm cal}$ is selected on the calibration prefix and
frozen for all held-out frames:
\begin{equation}
\text{reject as map-out if }q_t<\tau_{\rm cal};\qquad
\text{otherwise accept }\hat j_t^{\rm SVD}.
\label{eq:route-mht-gate}
\end{equation}
This decision changes no map index and uses no query coordinate or support
label at inference. The retained-map-hole experiment makes the acceptance
event observable without conflating it with closed-set Recall@1.

\paragraph{Parallel multimodal MHT tokens.}
Parallel modality tokenizers let attention combine visual, geometric, and motion evidence.
The image stream is a frozen AnyThermal--SALAD descriptor
$\mathbf v_t\in\mathbb R^{8448}$ projected by a fixed, label-independent
Gaussian matrix to $\tilde{\mathbf v}_t\in\mathbb R^{128}$; the projection is
not trained on the evaluation labels. The geometry stream is the nine-element
SVD vector $\mathbf r_t$ above, and the motion stream is $\mathbf u_t$ in
Eq.~\eqref{eq:motion-token}. Each stream has its own MLP tokenizer,
LayerNorm, modality embedding, and time embedding. For a causal window of
$W=16$, the resulting tokens are processed by multi-head attention with a
block-causal mask. The evaluated pooled model further decomposes the visual
descriptor into five query--candidate relation tokens and attends jointly to
the geometric baseline cost, residual, and motion tokens. A bidirectional cross-attention
variant lets visual states query geometry/motion and lets the resulting context
query visual states in return. The last-frame modality states are mean-pooled
by a map-out decision head. The completed channel-removal runs preserve the
SVD solver and scalar prior; held-out results are reported in the Results section. An
image-token width audit over
$\{32,64,128,256\}$ is included in the supplementary artifact; the primary
archived result uses width 64.

\paragraph{Training procedure and hyperparameters.}
The guardrail is trained independently within each scenario-held-out fold.
The first 500 frames of each scenario provide calibration-only
standardization and threshold selection; training uses the 13,000 windows
from the 13 training scenarios, while three validation scenarios are used for
early stopping and four scenarios remain untouched until testing. We use a
one-layer Transformer encoder with token width 64, eight modality/time tokens,
four attention heads, feed-forward width 128, GELU activations, dropout 0.1,
AdamW with learning
rate $3\times10^{-4}$ and weight decay $10^{-3}$, batch size 256, and 45
epochs. Binary cross-entropy with logits is the training loss. The image
descriptor is projected from 8,448 to 128 dimensions by a fixed Gaussian
matrix (seed 20260724); model seeds are fixed per fold and run. No test
scenario contributes to normalization, model selection, or gradient updates.

\paragraph{Deterministic track-hold recovery.}
Track-hold is a separate deterministic VPR recovery mechanism. Under map-hole contamination the route-local pool retains the
true position in only $61.5\%$ of held-out frames at $50\%$ contamination. While
$\max_j s_{t,j}$ stays below a calibration-frozen threshold, the SVD position is
coasted by heading-updated dead reckoning,
$\hat{x}_{t}=\hat{x}_{t-1}+\delta_t^{fwd}[\cos\theta_t,\sin\theta_t]$, with
$\theta_t=\theta_{t-1}+\delta_t^{yaw}$; when similarity recovers, the track
re-anchors to the SVD position. No per-frame oracle is used. Map-frame
alignment of the integrated relative yaw yields $+1.96$~pp macro R@1@5~m at
$50\%$ map-hole contamination
(Table~\ref{tab:track-hold}).

\subsection{Protected Selection and Auxiliary Controls}
Tensorized route-energy and pool-constrained continuous selectors are
auxiliary controls, not components of Route-MHT. Their objectives and
implementation details are documented in the supplementary material.

The final protected VPR outputs apply the same margin gate to both selectors. Let
$\tilde{j}^{z}_t$ denote the unprotected output for
$z\in\{\mathrm{app},\mathrm{Route\text{-}SVD baseline}\}$; the reported protected output is
\begin{equation}
\hat{j}^{z}_t=
\begin{cases}
\frontendpool_t[1], & s_t(\frontendpool_t[1])-s_t(\frontendpool_t[2])\geq\tau,\\
\tilde{j}^{z}_t, & \text{otherwise}.
\end{cases}
\label{eq:gate}
\end{equation}
where $\tau=0.0091$ is selected and frozen from calibration prefixes only; no
held-out frame or label is used to choose it. The gate is a conservative policy:
its held-out performance is reported directly in the results tables.

\subsection{Frozen-List SVD baseline Control}
The frozen-list SVD control separates geometric selection from candidate expansion.
\emph{Raw-Top-$K$ SVD} applies the
identical 12-frame geometric baseline branch score to all frozen frontend Top-100 candidates
and returns the minimum-cost branch; it never invokes Eq.~\eqref{eq:pool}.
This control is intentionally route-dependent and is not expected to dominate
the route-local system. Its role is to prevent re-query coverage gains from
being attributed to geometry.
\section{Evaluation Protocol}
\label{sec:evaluation-protocol}
\subsection{Dataset-A: Real-Robot Indoor Thermal Recordings}
\label{sec:dataset-a}
Dataset-A is our in-house collection of thermal videos acquired with a physical
mobile robot in a building interior with corridors, doorways, and furnished
spaces (Figure~\ref{fig:qualitative-retrieval-dataset_a}). The institution and collection site are
withheld for double-blind review. Six recorded traversals from two days form
the five localization pairs in Table~\ref{tab:data}: three same-day pairs share
map traversal 01, and two cross-day pairs reuse queries 02b and 03 with
previous-day maps. These are repeated traversals, not five independent sites.

The evaluated inputs are $640\times400$ grayscale thermal frames after video
cropping, sampled every second frame (0.08~s nominal spacing), with timestamps
and associated planar position/yaw records. The three map and three query
traversals contain 18,062 distinct sampled frames. Each query's first half is
reserved for calibration and its second half for evaluation; Table~\ref{tab:data}
reports the exact counts. Dataset-A evaluates localization and candidate
coverage; the transformer is trained on the Dataset-B scenario folds below.
Relative forward/turn increments derive from the pose records, not independent
odometry; query absolute coordinates are excluded from inference.

\begin{table}[t]
\caption{Dataset-A frame counts after sampling. Cal.: query calibration prefix;
Test: held-out suffix. Cross-day R2/R3 reuse the 02b/03 queries.}
\label{tab:data}
\centering
\small
\begin{tabular}{@{}lrrrr@{}}
\toprule
Pair & Map & Query & Cal. & Test \\
\midrule
01$\rightarrow$03 & 3,080 & 2,906 & 1,453 & 1,453 \\
01$\rightarrow$02b & 3,080 & 3,046 & 1,523 & 1,523 \\
01$\rightarrow$06 & 3,080 & 2,904 & 1,452 & 1,452 \\
Cross-day R2 & 3,127 & 3,046 & 1,523 & 1,523 \\
Cross-day R3 & 2,999 & 2,906 & 1,453 & 1,453 \\
\bottomrule
\end{tabular}
\end{table}

\subsection{Metrics}
Evaluation separates candidate coverage, selection efficiency, and final recall.
For a method $m$, its actual candidate pool $\mathcal{P}_m(t)$, selected map
index $\hat{j}_m(t)$, and radius $r$, we report
\begin{align}
C_m(r)&=\frac{1}{T}\sum_t \mathbb{1}\left[\min_{j\in\mathcal{P}_m(t)}
\|\mathbf{x}_j-\mathbf{q}_t\|_2<r\right],\\
R_m(r)&=\frac{1}{T}\sum_t \mathbb{1}\left[
\|\mathbf{x}_{\hat{j}_m(t)}-\mathbf{q}_t\|_2<r\right],\\
E_m(r)&=R_m(r)/C_m(r).
\end{align}
Here $C_m$, $R_m$, and $E_m$ denote coverage, Recall@1, and selection efficiency.
Query coordinates $\mathbf{q}_t$ serve only support labels and final metrics;
the relative-motion provenance is specified above. We report held-out
1/3/5/10~m R@1 with frozen prefix calibration and paired moving-block
bootstrap intervals (block lengths 20, 100, and 200).

Direct-VPR comparisons hold the data, frontend, and scoring protocol fixed.
All rows share frozen AnyThermal descriptors, map/query frames and ordering,
relative-motion streams, evaluation coordinates, radii, and Top-100 lists.
Only causal route-local re-query (12~m, at most 50 candidates) and selection
vary in Table~\ref{tab:r1}; labels never construct the pool.

\subsection{Baselines and Scope}
The baselines isolate retrieval, geometric verification, and same-pool selection.
The three primary controls are (i) frozen AnyThermal Top-1, (ii) Raw-Top-$K$
SVD baseline, which refines only the frozen Top-100 list, and (iii) a
confidence-preserving appearance-only selector in the exact route-local pool.
As an external classical sequence baseline, descriptor-adapted SeqSLAM searches
the full map using only frozen appearance similarities and historical query
rows; it does not receive relative controls or the route-local candidate pool.
The proposed method is the protected route-re-query-plus-SVD baseline system in
Eq.~\eqref{eq:gate}; its comparison with (iii) uses the same frozen margin
threshold and the same Top-1-plus-local-pool interface. Public benchmark
transfer checks extend evaluation beyond Dataset-A.

The MS2 transfer check uses official synchronized day/night thermal sequences.
Labels and descriptors were regenerated from raw
thermal-left frames. The map is sequence \texttt{2021-08-13-16-08-46} (2,543
frames) and the query is \texttt{2021-08-13-21-58-13} (2,539 frames). The
first 1,000 query frames are calibration-only; all reported MS2 numbers use
the remaining 1,539 query frames. AnyThermal-SALAD descriptors are extracted
once, before scoring, and the same descriptors, map, split, and 5/10~m radii
are used for every comparison. The released thermal-camera odometry supplies
a pose-source-derived motion proxy, not independent-odometry evidence.

\subsection{Map-Out Guardrail Protocol}
Map-out evaluation separates scenario-held-out learning from prefix calibration.
Dataset-B is a bank of 20 map-hole scenarios derived from public
STheReO--KAIST morning/evening recordings~\cite{yun2022sthereo}, not a new collection.
Each performance-blind scenario has three non-overlapping retained-map holes and at least
25 in-map and map-out frames in both calibration and held-out suffixes. The
generator continues sampling after invalid single-class splits, and every hole
range is asserted against the archived SVD baseline report. The first 500 frames select
normalization and the operating threshold; the remaining 500 frames are used
only for deterministic evaluation. Route-MHT instead uses five disjoint
scenario folds: 13 complete scenarios (13,000 causal windows) train the model,
three select early stopping, and four are held out for test. The five test
folds partition all 20 scenarios, and all results average three fixed seeds.
A test scenario contributes no frame to training, normalization, or stopping.

Support labels are constructed from query coordinates after retrieval and never
enter runtime tokens. Route-MHT is compared with the deterministic SVD
baseline; operating thresholds remain calibration-frozen within each test
scenario. The motion signal is the packaged causal relative-motion stream. We report
calibration-frozen FPR, AUC, recall, scenario-paired bootstrap
intervals, and the contamination breakdown. The held-out folds measure
scenario-level map-out generalization; the public benchmark checks provide
complementary cross-environment evidence.

\subsection{Held-Out Reproducibility Protocol}
The supplementary artifact freezes scenario partitions, calibration boundaries,
model seeds, and retained-map holes for reproduction of the protocol above.
\section{Results}
\subsection{Protected VPR accuracy}
\begin{figure*}[t]
\centering
\includegraphics[width=\textwidth]{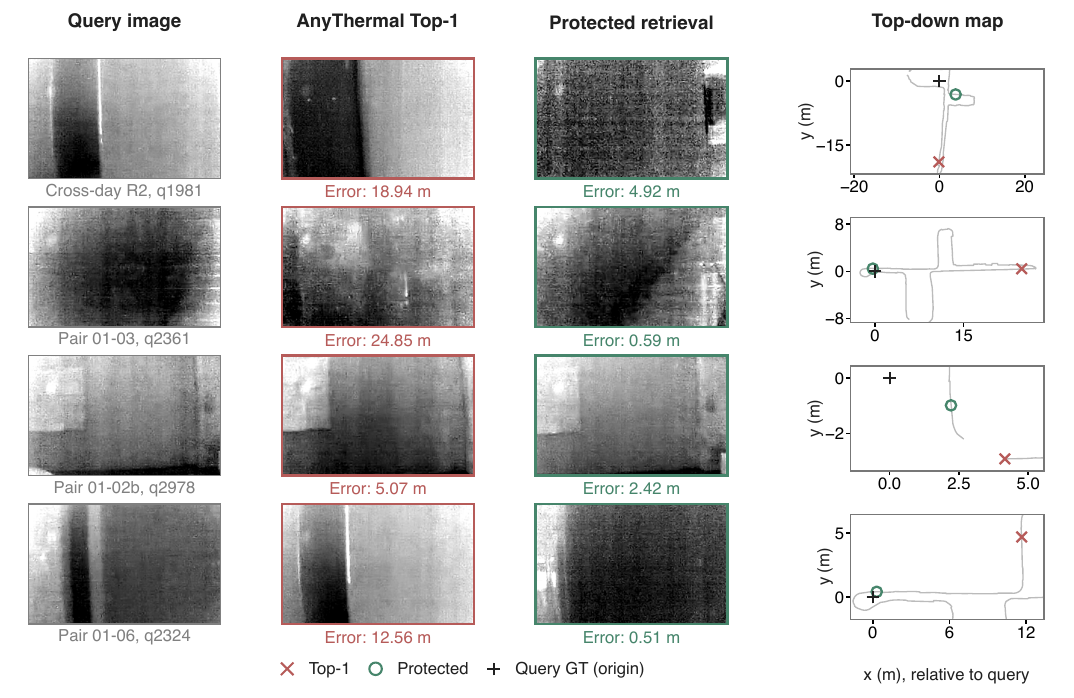}
\caption{Four selected Dataset-A retrieval cases: query, frozen AnyThermal
Top-1, protected retrieval, and top-down map. All images are displayed in
grayscale with per-image 2--98\% contrast limits. Prediction borders indicate
errors above (red) or within (green) 5~m. Map coordinates are translated so
the query ground truth is the origin; distances and aspect ratio are preserved.
These are supported-map retrieval examples, not MHT rejection outcomes.
Error and ground-truth annotations are evaluation-only. Aggregate recall is in
Table~\ref{tab:r1}.}
\label{fig:qualitative-retrieval-dataset_a}
\end{figure*}

Protected route-local SVD improves macro localization recall on five Dataset-A
pairs (Table~\ref{tab:r1}). The protected system uses the same frozen Top-100 frontend
list as the controls, opens a causal route-local pool, and applies SVD
verification with a calibration-frozen confidence rule.  The protected macro
R@1 values are $.610$ at 5~m and $.861$ at 10~m; these protected values are
not changed by any of the guardrail audits. Figure~\ref{fig:qualitative-retrieval-dataset_a}
shows four retrieval examples; Table~\ref{tab:vpr-reliability} summarizes
localization and map-out failure rates separately.

\begin{table}[t]
\caption{Dataset-A held-out Recall@1 (higher is better). Top-1: frozen
retrieval; Raw SVD: Top-100 verification; G-App.: same-pool appearance;
P: protected route-local SVD. Bold identifies P, not the row maximum.}
\label{tab:r1}
\centering
\footnotesize
\resizebox{\columnwidth}{!}{%
\begin{tabular}{@{}lrrrr@{}}
\toprule
Pair & Top-1 & Raw SVD & G-App. & \textbf{P} \\
\midrule
\multicolumn{5}{c}{R@1 $<5$ m} \\
01$\rightarrow$03 & .915 & .962 & .923 & \textbf{.937} \\
01$\rightarrow$02b & .139 & .156 & .139 & \textbf{.150} \\
01$\rightarrow$06 & .992 & .991 & .992 & \textbf{.995} \\
Cross-day R2 & .074 & .074 & .073 & \textbf{.076} \\
Cross-day R3 & .872 & .849 & .893 & \textbf{.891} \\
\midrule
\multicolumn{5}{c}{R@1 $<10$ m} \\
01$\rightarrow$03 & .970 & .991 & .977 & \textbf{.982} \\
01$\rightarrow$02b & .478 & .500 & .474 & \textbf{.499} \\
01$\rightarrow$06 & .994 & .998 & .994 & \textbf{.997} \\
Cross-day R2 & .874 & .906 & .885 & \textbf{.892} \\
Cross-day R3 & .928 & .912 & .944 & \textbf{.937} \\
\bottomrule
\end{tabular}
}%
\end{table}

The $R=C\cdot E$ decomposition separates candidate coverage from within-pool
selection efficiency.  Route-local re-query changes coverage without using
labels, while SVD changes only the selection within that same pool.  This is
important on the hard pairs: if the true position is absent from the pool,
no downstream selector can recover it.

\begin{table}[t]
\caption{Protected VPR and map-out guardrail summary (higher R@1, lower failure
and FPR). Closed-set failure is $1-\mathrm{R@1}$, whereas guardrail FPR is
measured separately on retained-map holes with explicit in-map/map-out labels.}
\label{tab:vpr-reliability}
\centering
\small
\setlength{\tabcolsep}{3pt}
\begin{tabular}{@{}lrrrrr@{}}
\toprule
& \multicolumn{2}{c}{@5 m} & \multicolumn{2}{c}{@10 m} & Guardrail \\
\cmidrule(lr){2-3}\cmidrule(lr){4-5}\cmidrule(lr){6-6}
Method & R@1 & Failure$\downarrow$ & R@1 & Failure$\downarrow$ & FPR$\downarrow$ \\
\midrule
Top-1 & .598 & .402 & .849 & .151 & .366 \\
Protected SVD & .610 & .390 & .861 & .139 & .116 \\
\bottomrule
\end{tabular}
\end{table}

\subsection{Trajectory-consistent track-hold}
Track-hold freezes the threshold on the calibration prefix, coasts through a
low-confidence interval with the packaged relative-motion stream, and
re-anchors when a valid SVD match returns.  It is deterministic and separate
from any learned selector.

\begin{table}[t]
\caption{Track-hold stress test over five Dataset-A pairs. Entries are macro
R@1@5~m; $\Delta$ is the percentage-point change from SVD. Contamination
uses SHA256-stable retained-map holes.}
\label{tab:track-hold}
\centering
\small
\begin{tabular}{@{}lrrrr@{}}
\toprule
Contamination & 0\% & 20\% & 50\% & 70\% \\
\midrule
SVD baseline & .604 & .576 & .498 & .415 \\
Track-hold & .610 & .610 & .518 & .408 \\
$\Delta$ (pp) & +.60 & +3.34 & +1.96 & -.71 \\
\bottomrule
\end{tabular}
\end{table}

Track-hold improves moderate-contamination recovery but not the most severe case.
These values align integrated relative yaw to the ordered map tangent at each
visual anchor. The positive gain through 50\% does not
extend to 70\%. Cross-day R2 remains the limiting pair because its
dead-reckoning error exceeds 16~m; this failure is retained as the operating
boundary rather than tuned away.

\subsection{Public benchmark transfer checks}
We additionally audit the frozen route-local SVD interface on three public
thermal benchmarks. The same AnyThermal descriptor source and held-out scoring
code are used for each dataset; Table~\ref{tab:external-vpr} reports direct retrieval rather than
a claim that the multimodal guardrail was retrained on every benchmark.
STheReO--KAIST \cite{yun2022sthereo} provides metric pose products, MS2
\cite{shin2025ms2} uses a newly downloaded day/night synchronized split, and
IRSLAM--KRI \cite{keil2024irslam} uses a route-progress proxy
rather than metric ground truth. These distinctions are retained so that
cross-environment transfer is not confused with the scenario-held-out
map-out audit.

\begin{table}[t]
\caption{Public held-out R@1 at 5/10~m. All methods use frozen descriptors and
fixed suffixes. MS2 uses 1,539 held-out queries after 1,000 calibration frames;
MS2 and IRSLAM--KRI use pose-derived or route-progress proxies, not independent
odometry. Each cell is R@1@5~m / R@1@10~m. MS2 uses a 5,082-frame day/night
run. ``Protected'' denotes the proposed route-local interface.}
\label{tab:external-vpr}
\centering
\small
\setlength{\tabcolsep}{3pt}
\begin{tabular}{@{}lccc@{}}
\toprule
Method & \shortstack{STheReO\\KAIST} & MS2 & \shortstack{IRSLAM\\KRI} \\
\midrule
AnyThermal Top-1 & .038/.730 & .301/.606 & .018/.032 \\
Causal reranking & .010/.934 & .327/.695 & .016/.024 \\
Temporal posterior & .024/.876 & .305/.634 & .006/.006 \\
Raw Top-$K$ geometric & .050/.766 & .366/.682 & .040/.046 \\
Route-local appearance & .056/.802 & .327/.672 & .022/.024 \\
Protected appearance & .050/.772 & .309/.643 & .018/.024 \\
Route + geometry & .056/.804 & .356/.689 & .024/.024 \\
\textbf{Protected SVD} & \textbf{.054/.790} & \textbf{.327/.654} & \textbf{.020/.024} \\
SeqSLAM & .138/.734 & .128/.244 & .032/.050 \\
\bottomrule
\end{tabular}
\end{table}

\subsection{Multimodal MHT operating-point audit}
Route-MHT improves the calibrated map-out operating point over the geometric
baseline (Table~\ref{tab:multimodal-mht-audit}). Evaluation uses five scenario-held-out folds,
13,000 training windows per fold, three validation scenarios, four test
scenarios, and three seeds. Across 20 test scenarios, Route-MHT obtains
FPR/AUC/recall $=.061/.967/.914$, compared with $.116/.945/.884$ for the SVD baseline.
Paired scenario bootstrap intervals for MHT minus SVD baseline are
$[-.100,-.014]$ for FPR, $[+.015,+.029]$ for AUC, and
$[+.019,+.041]$ for recall. These paired results show that the learned
multimodal decision improves the calibration-frozen map-out operating point
without changing the SVD pose solver or candidate index.

Reproducible profiling bounds backend cost on an Apple M4 MacBook Air (10-core
CPU, 16~GB; Python~3.13.13, PyTorch~2.12.0). Over ten runs of 1,000
STheReO--KAIST queries, route expansion plus SVD takes
$3.443\!\pm\!0.055$~ms/frame (median/P95: 3.436/3.522). The 90,817-parameter
MHT takes 0.441/0.491~ms median/P95 on CPU (FP32, batch one, four threads,
200 warm-ups, 2,000 trials). These cached-descriptor timings exclude decoding,
AnyThermal extraction, sensor I/O, and process transfer; they are backend-only,
not sensor-to-decision latency.

Channel removal identifies visual and motion evidence as rejection cues
(Table~\ref{tab:channel-ablation}; identical folds and three seeds). Removing
image or motion tokens raises FPR to $.086$ or $.074$, respectively; removing
geometry diagnostics leaves FPR at $.061$ while recall changes from $.914$ to
$.920$. The fixed geometric solver and scalar prior remain active throughout.

\begin{table}[t]
\caption{Map-out guardrail audit on 20 frozen scenarios with calibration-fixed
thresholds; this diagnostic is separate from protected VPR.}
\label{tab:multimodal-mht-audit}
\centering
\small
\begin{tabular}{@{}lrrr@{}}
\toprule
Method & FPR$\downarrow$ & AUC$\uparrow$ & Recall$\uparrow$ \\
\midrule
Geometric baseline & .116 & .945 & .884 \\
Route-MHT & \textbf{.061} & \textbf{.967} & \textbf{.914} \\
\bottomrule
\end{tabular}
\end{table}

\begin{table}[t]
\caption{Completed leave-one-channel-out ablation over five folds and three
seeds. The SVD pose solver and scalar safety prior remain unchanged.}
\label{tab:channel-ablation}
\centering
\small
\begin{tabular}{@{}lrrr@{}}
\toprule
Variant & FPR$\downarrow$ & AUC$\uparrow$ & Recall$\uparrow$ \\
\midrule
Full Route-MHT & .061 & .967 & .914 \\
w/o Image & .086 & .961 & .911 \\
w/o Geometry diagnostics & .061 & .967 & .920 \\
w/o Motion & .074 & .967 & .919 \\
\bottomrule
\end{tabular}
\end{table}
\section{Discussion}
\subsection{What the evidence supports}
The evidence supports Route-MHT as a multimodal trust layer: route-local
re-query changes coverage, SVD verifies geometry, and attention audits map
support. Its held-out FPR improvement has a paired interval excluding zero,
while the protected five-pair interface retains its reported recall.

\subsection{Candidate coverage and temporal recovery}
The $R=C\cdot E$ decomposition separates re-query coverage from selection.
Neither stage recovers locations absent from the pool: at 50\% contamination,
coverage is 61.5\%, which bounds recall regardless of the selector.

Track-hold bridges temporary confidence drops using pose-derived motion. It
adds 1.96~pp macro R@1@5~m at 50\% contamination, but reverses at 70\%; without
an oracle reset, cross-day R2 exceeds 16~m dead-reckoning error.

\subsection{Guardrail operating point}
Ranking quality and calibrated false acceptance are distinct: AUC measures
ranking, whereas the retained-map-hole operating point measures false
acceptance. The SVD baseline's large FPR change despite a small AUC change
therefore motivates reporting both metrics.

\subsection{Operating boundary}
The evidence is bounded by pose-derived motion, controlled map holes, and
source-specific public protocols. Deployment requires independent sensor
validation and additional untouched routes.
\section{Conclusion}
This paper presents \methodname{}, a thermal-navigation guardrail combining a
causal route-local interface, exact $SE(2)$ verification, and masked attention
over visual, residual, and motion tokens without replacing the SVD pose solver.
The protected interface reaches macro R@1 $.610/.861$ at 5/10~m; across 20
scenario-held-out map holes, \methodname{} lowers FPR $.116\!\rightarrow\!.061$
while raising AUC $.945\!\rightarrow\!.967$ and recall
$.884\!\rightarrow\!.914$. STheReO--KAIST, MS2, and IRSLAM--KRI provide public
transfer checks on robot-collected thermal traversals. Future work will extend the guardrail to multi-robot mapping
and large-scale outdoor navigation.
\FloatBarrier

\end{document}